\documentclass{article}
\usepackage[T1]{fontenc}
\usepackage{iclr2026_conference,times}
\iclrfinalcopy

\usepackage{graphicx}
\usepackage{xcolor}
\usepackage{geometry}
\usepackage{amsmath}
\usepackage{amssymb}
\usepackage{booktabs}
\usepackage{multirow}
\usepackage{array}
\usepackage{colortbl}
\usepackage{enumitem}
\usepackage[most]{tcolorbox}
\usepackage{tikz}
\usetikzlibrary{arrows.meta,positioning,fit,calc,backgrounds}
\usepackage{url}
\usepackage{fancyhdr}

\newcommand{\dz}{\ensuremath{\Delta_{\mathrm{zero}}}}
\newcommand{\ds}{\ensuremath{\Delta_{\mathrm{same}}}}
\newcommand{\dpa}{\ensuremath{\Delta_{p}}}
\newcommand{\obs}{\ensuremath{\mathrm{obs}}}
\newcommand{\Rsite}{\ensuremath{R_{\mathrm{site}}}}
\newcommand{\hi}{\ensuremath{\mathrm{hi}}}
\newcommand{\lo}{\ensuremath{\mathrm{lo}}}
\newcommand{\worlds}{\mathcal{W}}

\title{Located but Not Releasable: Silent Gate Inversion and Bounded Linear Release}

\author{Xining Xun\\
Tsingjiao Information Science (Beijing) Co., Ltd.}

\date{}

\begin{document}
\maketitle

\pagestyle{fancy}
\fancyhf{}
\lhead{Preprint}
\rhead{\thepage}
\renewcommand{\headrulewidth}{0.4pt}

\begin{abstract}
A growing body of work reports that language models \emph{represent} task-relevant latent structure that they fail to \emph{use}. Whether such structure, once located, can be converted into behavior is a separate question that is rarely tested end to end. We submit the complete pipeline---detect, localize, and release---to a fully preregistered stress test on a 25.7M transformer trained on causal-evidence discrimination, where a known \emph{suppression} phenomenon (latent causal structure present but behaviorally unused) has previously been documented. Every threshold, every claim template, and every branch of the decision tree was hashed and archived before any corresponding data existed. Three findings emerge. (i)~\textbf{Localization succeeds}: interventions at observation-evidence channels of mid layers restore target behavior on otherwise-suppressed worlds (paired release advantage $0.563$ and $0.854$, 97.5\% CIs excluding zero; best-site release rate $0.889$). (ii)~\textbf{Gating fails out of distribution}: a detector calibrated to trigger on zero out-of-distribution calibration worlds triggers on 6.9--7.3\% of held-out in-distribution generations and on \emph{zero} of the 2{,}400 held-out generations that actually need it---a complete inversion that silently reduces the gated pipeline to its base model. (iii)~\textbf{Linear release is capped}: removing the gate entirely and injecting a per-instance linear direction unconditionally yields a monotone dose--response that plateaus far below the preregistered release margin (intercept $0.382 \!\to\! 0.311 \!\to\! 0.264$ vs.\ threshold $\le 0.08$), and per-instance adaptivity adds less than $\pm 0.03$ over the fixed direction. The failure is therefore doubly located: the detector is OOD-inverted, \emph{and} the entire family of linear release directions at this site and resolution is bounded away from sufficiency. The two failures are dissociable, and neither overturns the localization result---representation-level localization and behavior-level release come apart. All artifacts, hashes, and claim templates are released for audit.
\end{abstract}

\vspace{0.5em}

\section{Introduction}\label{sec:intro}

Interpretability research has become very good at \emph{finding} structure: probes recover latent variables from activations~\citep{belinkov2022probing,hewitt2019designing,burns2023discovering}, causal-tracing methods localize the components that mediate a behavior~\citep{vig2020investigating,meng2022locating}, and steering vectors shift model outputs along semantically meaningful directions~\citep{turner2023activation,zou2023representation,li2023inference,subramani2022extracting}. Implicit in much of this literature is a pipeline assumption: once structure is \emph{detected} and \emph{localized}, \emph{releasing} it into behavior is an engineering matter.

Recent evidence complicates this assumption. Steering vectors generalize unreliably across prompts and distribution shifts~\citep{tan2024analysing}; strong detectors of internal states coexist with failed control in concurrent work~\citep{geometrysteering2026}; and in recent work, a small transformer trained on causal-evidence discrimination was shown to \emph{suppress}---rather than lack---causal structure: the structure is linearly decodable from its residual stream, yet the model behaves as if it were absent~\citep{xun2026evidence}. Together these results suggest a representation--behavior gap that is structural, not incidental. What they do not answer is the question that matters for intervention-based methods: \textbf{if we locate the suppressed structure precisely, can we release it into behavior, end to end, under preregistered criteria?}

This paper answers that question with a complete, preregistered failure decomposition (Figure~\ref{fig:pipeline}). We take the suppression phenomenon of \citet{xun2026evidence} as ground truth and build a four-component pipeline---an OOD detector $u$, a contrast forward pass, a direction generator (calibration-fitted $g_\theta$ or per-instance), and a residual-stream injector---that must (a) decide when intervention is needed, (b) construct the intervention, and (c) deliver it without damaging in-distribution (ID) behavior. Every threshold, evaluation set, statistic, claim template, and fallback branch was fixed and hashed \emph{before} any corresponding measurement (Section~\ref{sec:setup}); the full audit chain is reproduced in Appendix~\ref{app:audit}.

Our findings decompose the failure into two dissociable parts:

\begin{enumerate}[leftmargin=1.6em,itemsep=0.2em]
  \item \textbf{Localization is real and sufficient.} Transplanting activations at observation-evidence channels of mid layers restores target behavior on suppressed worlds, with paired advantages $0.563$ and $0.854$ (97.5\% confidence intervals excluding zero) and a best-site release rate of $0.889$ (Section~\ref{sec:local}). The structure to be released exists and is site-specific.
  \item \textbf{End-to-end release fails twice, for two independent reasons.} First, the detector that passed its calibration with \emph{zero} OOD triggers inverts out of distribution: it fires on 6.9--7.3\% of held-out ID generations and on exactly 0 of 2{,}400 generations from the worlds that need intervention, so the ``gated'' pipeline is behaviorally identical to the base model (Section~\ref{sec:gate}). Second, even with the gate removed entirely---unconditional injection of a per-instance linear direction at the localized site---behavior moves monotonically with dose but plateaus far below the preregistered release margin; halving the dose loses performance monotonically, and per-instance adaptivity contributes at most $\pm 0.03$ over a fixed direction (Section~\ref{sec:release}). The entire \emph{family} of linear release directions at this site and resolution is thus bounded away from sufficiency.
\end{enumerate}

The decomposition is the contribution. Neither failure alone would license our conclusions: an OOD-inverted detector leaves open the possibility that a better detector rescues release; a capped linear direction leaves open the possibility that the gate was the only obstacle. Together, and only together with the localization result, they establish that the representation--behavior gap in this system is not a localization gap, not (only) a routing gap, and not closable by any linear release direction at the tested resolution---a conclusion that preregistration protects from post-hoc re-narration. We discuss implications for steering-vector practice and for evaluation standards in Section~\ref{sec:discussion}.

\paragraph{Preregistration and reproducibility.} All criteria, thresholds, seeds, claim templates, and decision-tree branches were archived with MD5 hashes before the corresponding data existed; every number reported in this paper traces to a hashed artifact, and the release includes the audit log, code manifests, and per-world records (Appendix~\ref{app:audit}). Total measured cost of the terminal experiments reported here is $4.22$ GPU-hours on a single consumer-class GPU.

\section{Related Work}\label{sec:related}

\paragraph{Probing and the limits of representation-level claims.} Linear and nonlinear probes recover a wide range of latent variables from activations~\citep{belinkov2022probing}, but probe accuracy is not evidence of use: control tasks and information-theoretic bounds show that decodability and functional role dissociate~\citep{hewitt2019designing}. Unsupervised methods can extract latent knowledge without labels~\citep{burns2023discovering}. Our starting point accepts this critique: stage-1 of our audit (Section~\ref{sec:stage1}) itself fails a purely probe-based account, which is precisely why the subsequent stages test \emph{interventions} rather than decodings.

\paragraph{Activation steering and representation engineering.} Adding direction vectors to hidden states can steer sentiment, style, truthfulness, and refusal~\citep{subramani2022extracting,turner2023activation,zou2023representation,li2023inference,rimsky2024steering,arditi2024refusal}. However, steering vectors generalize unevenly across prompts and are sensitive to distribution shift~\citep{tan2024analysing}, and single-direction mediation results are typically reported at the level of average effects rather than preregistered behavioral criteria. Our B6.6 result quantifies one such boundary precisely: at a \emph{verified} causal site, the linear-direction family has a measurable, monotone, and insufficient dose--response curve.

\paragraph{Causal mediation and localization.} Causal tracing and interchange interventions localize mediators of factual recall, gender bias, and syntactic agreement~\citep{vig2020investigating,meng2022locating,geiger2022inducing}. We use a transplant intervention in this tradition for our localization stage (Section~\ref{sec:local}), but unlike most localization work we carry the located site forward into an end-to-end behavioral pipeline and let it fail publicly under frozen criteria.

\paragraph{Causal reasoning in language models.} CLadder~\citep{jin2023cladder} evaluates formal causal reasoning across the ladder of causation~\citep{pearl2009causality}; recent work documented suppression of latent causal structure in a small transformer trained on causal-evidence discrimination~\citep{xun2026evidence}. The present paper closes the loop: it tests whether the documented-but-suppressed structure can be released into behavior, and decomposes the failure.

\paragraph{Concurrent work.} The independent concurrent study \emph{The Geometry of Knowing vs.\ Steering}~\citep{geometrysteering2026} reports ``perfect detection, failed control'' in large language models, conceptually convergent with our detector--release dissociation. Our results differ in method (preregistered decision trees with hashed claim templates, a controlled synthetic causal domain, and a localized causal site) and in conclusion granularity: we separate \emph{gate} failure (OOD-inverted detector) from \emph{release} failure (bounded linear direction family) and show each is independently sufficient to sink the pipeline.

\paragraph{Preregistration and negative results.} Undisclosed analytic flexibility inflates false positives~\citep{simmons2011false}; preregistration constrains it~\citep{nosek2018preregistration}. Machine-learning evaluations rarely preregister \emph{claim templates}---verbatim sentences the authors commit to publishing conditional on outcomes. We do, and we treat both failed terminal experiments as reportable results rather than pilots.

\section{Setup, Measurements, and Preregistration Protocol}\label{sec:setup}

\subsection{Model and task}

We study \textbf{C1}, a $25.7$M-parameter GPT-style decoder-only transformer~\citep{vaswani2017attention} (8 layers, na\"ive attention, vocabulary 41) trained on causal-evidence discrimination as specified in \citet{xun2026evidence}: prompts present observational evidence under a structural causal model and the model answers interventional queries $\mathrm{do}(X{=}x_0) \to Y{=}?$. The training distribution contains \emph{Simpson-trap} worlds in which the observational correlation and the causal effect have opposite signs. The suppression phenomenon of interest: on held-out confounded worlds the model's behavior tracks the observational association even though the causal structure is linearly decodable from its residual stream~\citep{xun2026evidence}. We use the frozen checkpoint of \citet{xun2026evidence} throughout; no parameter of C1 is ever updated in this paper.

\subsection{Behavioral measurements}\label{sec:meas}

All behavioral statistics derive from a two-point response grid. For a world $w$ with answer-relevant span $[\lo, \hi]$ we measure the unit-span slope of the model's parsed numeric response,
\begin{equation}
  \Delta \;=\; \frac{\mathrm{parse}(r(\hi)) - \mathrm{parse}(r(\lo))}{\hi - \lo},
  \label{eq:delta}
\end{equation}
where $r(\cdot)$ is greedy decoding and $\mathrm{parse}$ extracts the signed numeric answer (unparseable outputs are rejected conservatively). Each world supplies three slope measurements:
\begin{align}
  \ds &= \Delta \text{ under the same evidence}; \notag\\
  \dz &= \Delta \text{ under evidence zeroed at \obs\ channels}; \label{eq:three}\\
  \dpa &= \Delta \text{ under intervention}. \notag
\end{align}
The \emph{denominator} $D = \dz - \ds$ quantifies how much behavior the zeroed evidence removes; a world is \textbf{measurable} iff $|D| \geq 0.05$ (the denominator guard; unmeasurable worlds are excluded from rate statistics but retained in the audit). The \emph{release rate} of an intervention is
\begin{equation}
  R \;=\; \frac{\dpa - \ds}{\dz - \ds} \;\in\; (-\infty, \infty),
  \label{eq:R}
\end{equation}
the fraction of the zeroed-evidence behavioral shift restored by the intervention. $R{=}1$ is perfect release; $R{=}0$ is no effect; $R{<}0$ overshoots in the wrong direction.

For population-level release we fit the \emph{offset-gain} regression over a world set $\worlds$,
\begin{equation}
  \widehat{\Delta_w} \;=\; a \cdot \mathrm{ATE}_w + b, \qquad w \in \worlds,
  \label{eq:fit}
\end{equation}
where $\mathrm{ATE}_w$ is the world's ground-truth causal effect. The ideal responder has $a{=}1, b{=}0$; a suppressed model exhibits large $b$ (a constant pull toward the observational answer). Our preregistered release criteria read off the \textbf{intercept} $b$ (must drop to $\le +0.08$) and require the ID slope $a$ to not degrade by more than $0.05$.

Two supporting statistics recur. \emph{Reversal count}: on worlds with $\mathrm{ATE}_w < 0$, the number of worlds on which the sign of the prediction disagrees with the sign of the ATE with margin $|{\cdot}| > 0.05$. \emph{McNemar exact test}~\citep{mcnemar1947note}: for paired before/after correctness, one-sided $p = \Pr[\mathrm{Bin}(b{+}c, \tfrac12) \le c]$ over discordant pairs $(b, c)$. Rates are reported with Wilson 95\% intervals~\citep{wilson1927probable}; regression intercepts and paired differences with bootstrap 95\% intervals~\citep{efron1993introduction}; localization confidence intervals are Bonferroni-corrected across the compared pairs.

\subsection{Representation-level measurements}

For representation comparisons we use centered linear CKA~\citep{kornblith2019similarity}, i.e.\ the RV coefficient
\begin{equation}
  \mathrm{CKA}(X, Y) \;=\; \frac{\mathrm{tr}(\Sigma_X \Sigma_Y)}{\sqrt{\mathrm{tr}(\Sigma_X^2)\,\mathrm{tr}(\Sigma_Y^2)}},
  \label{eq:cka}
\end{equation}
computed on mean-pooled, layer-normalized features. The site feature map is
\begin{equation}
  \phi_\ell(x) \;=\; \mathrm{LayerNorm}\big(\mathrm{MeanPool}_{\obs}\big(h^{(\ell)}(x)\big)\big),
  \label{eq:phi}
\end{equation}
mean-pooling over the observation-evidence token rows of the residual stream at layer $\ell$, concatenated across the site's layers. The detector score is a probe $u(x) \in \mathbb{R}$ trained to separate confounded from non-confounded worlds; the pipeline triggers intervention iff $u(x) \ge T$ for a threshold $T$ fixed at calibration.

\subsection{The CCG pipeline}

\begin{figure}[t]
\centering
\begin{tikzpicture}[
  font=\footnotesize,
  box/.style={draw, rounded corners=2pt, align=center, minimum height=8mm, inner sep=3pt},
  dec/.style={draw, diamond, aspect=2.2, align=center, inner sep=1.5pt, font=\footnotesize},
  arr/.style={-{Stealth[length=2.5mm]}, thick},
  lab/.style={font=\footnotesize, midway, above}
]
\node[box, fill=blue!6] (det) at (0,0) {\textbf{Detect}\\$u(x)$ vs $T$};
\node[box, fill=blue!6, right=6mm of det] (con) {\textbf{Contrast}\\$h_{\mathrm{zero}}$ vs $h_{\mathrm{same}}$};
\node[box, fill=blue!6, right=6mm of con] (gen) {\textbf{Generate}\\$\delta = g_\theta(\cdot)$ or Eq.~\eqref{eq:dx}};
\node[box, fill=blue!6, right=6mm of gen] (inj) {\textbf{Inject}\\$h \leftarrow h + \lambda\delta$};
\node[box, fill=green!10, right=6mm of inj] (out) {\textbf{Answer}\\parse + guard};
\node[box, fill=gray!10, below=8mm of det] (id) {ID branch (unmodified)};
\draw[arr] (det) -- (con);
\draw[arr] (con) -- (gen);
\draw[arr] (gen) -- (inj);
\draw[arr] (inj) -- (out);
\draw[arr] (det) -- node[lab, left, xshift=-1mm]{$u<T$} (id);
\draw[arr] (id.east) -| (out.south);
\begin{scope}[on background layer]
\node[draw=red!60, dashed, rounded corners=3pt, fit=(det), inner sep=3pt,
      label={[red!70!black, font=\footnotesize]below:{\shortstack{\textbf{Failure 1} (B6.5): OOD-inverted;\\0/2{,}400 triggers where needed}}}] {};
\node[draw=red!60, dashed, rounded corners=3pt, fit=(gen)(inj), inner sep=3pt,
      label={[red!70!black, font=\footnotesize]below:{\shortstack{\textbf{Failure 2} (B6.6): linear direction family\\bounded below release margin}}}] {};
\end{scope}
\end{tikzpicture}
\caption{The CCG release pipeline and the two failure locations established in this paper. Detection gates the entire intervention path; B6.5 shows the gate inverts out of distribution and never fires on the worlds that need it. B6.6 removes the gate (unconditional injection) and shows the linear release-direction family at the localized site still cannot reach the preregistered margin. The localization stage (Section~\ref{sec:local}) sits upstream of both and passes.}
\label{fig:pipeline}
\end{figure}
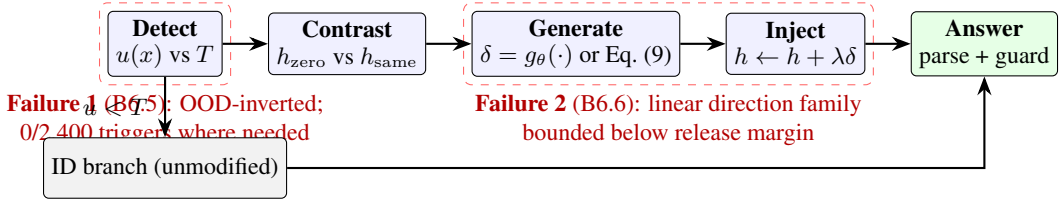

The \emph{contrastive causal gate} (CCG) inference pipeline has four components (Figure~\ref{fig:pipeline}): (i)~\textbf{detect}---score $u(x)$ and compare against $T$; if $u(x) < T$ the prompt follows the ID branch and is answered unmodified; (ii)~\textbf{contrast}---run a second forward pass with evidence zeroed at \obs\ channels; (iii)~\textbf{generate}---a small learned map $g_\theta$ produces a direction $\delta$ from the contrast pair; (iv)~\textbf{inject}---the residual stream is modified as
\begin{equation}
  h^{(\ell)}(x) \;\leftarrow\; h^{(\ell)}(x) + \lambda \cdot \delta(x), \qquad \ell \in \text{site},
  \label{eq:inject}
\end{equation}
followed by a second decoding; conservative refusal rules reject unparseable outputs and degenerate contrasts. Two direction constructions are studied. The calibration-fitted direction (B6.5) uses a two-segment mean-difference vector
\begin{equation}
  \delta_{\mathrm{ZERO}} \;=\; \overline{\mathrm{MeanPool}}\big(h_{\mathrm{zero}}\big) \;-\; \overline{\mathrm{MeanPool}}\big(h_{\mathrm{same}}\big),
  \label{eq:dzero}
\end{equation}
with \emph{raw} mean pooling (no LayerNorm), frozen after calibration. The per-instance direction (B6.6) removes all learned parameters and the gate:
\begin{equation}
  \delta(x) \;=\; \mathrm{MeanPool}_{\mathrm{raw}}\big(h_{\mathrm{zero}}(x)\big) - \mathrm{MeanPool}_{\mathrm{raw}}\big(h_{\mathrm{same}}(x)\big),
  \label{eq:dx}
\end{equation}
computed per prompt and injected unconditionally ($\lambda \in \{0.5, 1.0\}$) or always-on as a fixed control.

\subsection{Preregistration protocol}\label{sec:prereg}

Every experiment reported here follows the same discipline, archived in a single append-only audit log (Appendix~\ref{app:audit}):
\begin{enumerate}[leftmargin=1.6em,itemsep=0.15em]
  \item \textbf{Criteria before data.} The full criteria file---thresholds, evaluation sets, seeds, statistics, and verbatim claim templates---is hashed (MD5) and archived \emph{before} any corresponding measurement is produced.
  \item \textbf{Frozen branches.} Each outcome maps to a pre-written claim template; missed criteria follow the pre-written failure branch. Thresholds are never lowered, statistics never re-specified, and seeds never added after seeing data.
  \item \textbf{Protocol amendments precede data.} Any amendment (e.g.\ the B6.6 amendment of Section~\ref{sec:release}) is archived before any data it governs exists.
  \item \textbf{Pilot transparency.} Pilot results that informed design choices are reported alongside confirmatory results, never silently absorbed.
\end{enumerate}
The terminal experiments have fixed evaluation sets: \textbf{CONF50} (50 held-out confounded worlds at full power), \textbf{ID40} (40 held-out non-confounded worlds), and a \textbf{calibration} set of 40 worlds (45 drawn minus 5 overlapping CONF50) spanning 8 families with 9 positive worlds scattered across 7 families.

\section{Stage 1: Three Gates Before Any Release Attempt}\label{sec:stage1}

Before attempting release, the audit chain subjected the naive hope---``the structure is decodable, so release should follow''---to three preregistered gates (Table~\ref{tab:stage1}). All three failed. These failures are not the subject of this paper, but they define why the pipeline of Figure~\ref{fig:pipeline} was necessary at all, and they supply the baselines against which the later results are read.

\paragraph{B0: behavior-coupled readout does not exist.} Probes trained to predict the causal answer from residual features achieve measurable behavioral coupling only on $11/50$ worlds ($< 25$, the preregistered sentinel). On the measurable subset, mean release rates by site are $0.785$ ($[0.637, 0.932]$, L23$_{\obs}$), $0.829$ ($[0.663, 0.994]$, L345$_{\obs}$), $0.789$ ($[0.587, 0.962]$, L2345$_{\mathrm{all}}$), $0.326$ ($[0.112, 0.613]$, L23$_{\mathrm{non\text{-}obs}}$), and $0.186$ ($[{-0.104}, 0.461]$, L345$_{\mathrm{non\text{-}obs}}$): coupling concentrates on observation-evidence rows, but $39/50$ worlds fall below the denominator guard and are unmeasurable.

\paragraph{B0.5: representations do not align across evidence types.} Pairwise CKA (Eq.~\eqref{eq:cka}) between the confounded-world features and their observational counterparts ranges $0.061$--$0.068$ across 9 families, against a within-family reference scale with mean $0.431$ $[0.144, 0.998]$ and threshold $0.7$. The two evidence types occupy nearly orthogonal feature geometries, foreclosing ``readout transfer'' and motivating the gated-route design.

\paragraph{B1: the answer is detectable but not usable by the readout.} A $k{=}32$ MLP probe separates world families with AUC $0.962$, yet the behaviorally relevant signal is absent where it matters: ID trigger rate $0.100$ (at the boundary), worst leave-one-family-out AUC $0.366$ (ood-sign $0.640$, ood-scale $0.366$, ood-$\rho$ $0.658$), and zero-shot CLadder transfer AUC $0.022$--$0.292$ with \emph{reversed} direction---the probe's ranking anti-predicts external causal difficulty. A linear head reaches only $0.70$--$0.74$ where the MLP reaches $0.962$: the decodable information is nonlinearly encoded and does not survive distribution shift in the direction behavior would need.

\begin{table}[t]
\centering
\footnotesize
\caption{Stage-1 audit summary. Each gate was preregistered with its threshold and its failure branch; all three failed, and each failure dictated the design of the next stage. Intervals are 95\% (bootstrap or Wilson as marked in the audit).}
\label{tab:stage1}
\begin{tabular}{@{}ll>{\raggedright\arraybackslash}p{2.9cm}>{\raggedright\arraybackslash}p{3.4cm}l@{}}
\toprule
Gate & Question & Preregistered criterion & Result & Verdict \\
\midrule
B0 & Is readout behavior-coupled? & $\ge 25/50$ measurable worlds & $11/50$ & \cellcolor{red!12}FAIL \\
B0.5 & Do features align across evidence? & CKA $\ge 0.7$ vs ref.\ $0.431$ & $0.061$--$0.068$ & \cellcolor{red!12}FAIL \\
B1 & Is the signal usable by readout? & OOD/transfer AUC above floor & worst $0.366$; CLadder reversed & \cellcolor{red!12}FAIL \\
\bottomrule
\end{tabular}
\end{table}

Stage 1 therefore establishes the problem's shape: the causal structure is present (it is decodable), but it is neither behaviorally coupled, nor geometrically aligned across evidence types, nor usable by a generalizing readout. Release, if possible at all, must go through intervention on the residual stream---which first requires knowing \emph{where} to intervene.

\section{Localization Succeeds: The Suppression Is Site-Specific}\label{sec:local}

The localization stage (B0-rep) asks whether transplanting activations at a candidate site can restore suppressed behavior, under preregistered criteria C1 (paired release advantage over two control pairs, 97.5\% Bonferroni-corrected CIs excluding zero) and C2 (best-site release rate $\ge 0.5$). Sites are layer sets crossed with token-row classes (\obs\ vs non-\obs); the transplant replaces the site's residual rows of a suppressed world with those of its released counterpart.

\begin{figure}[t]
\centering
\includegraphics[width=\textwidth]{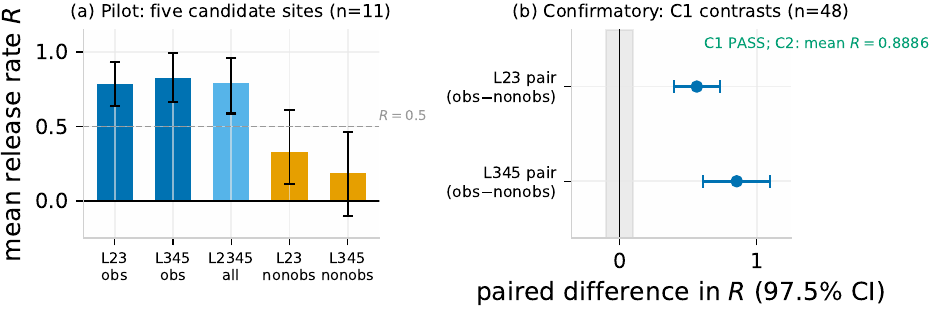}
\caption{Localization. \textbf{(a)} Pilot screen across five candidate sites (release rate, Eq.~\eqref{eq:R}); \obs-row sites dominate, and L23$_{\obs}$ is selected as the confirmatory site. \textbf{(b)} Confirmatory paired release advantages (Eq.~\eqref{eq:c1}) with 97.5\% Bonferroni-corrected intervals; both intervals exclude zero. Pilot numbers are archival and reported alongside confirmatory numbers per the preregistration protocol.}
\label{fig:local}
\end{figure}

\paragraph{Result: full pass.} Of $608$ screened worlds, $M = 48$ met the eligibility criteria (rate $0.0789$, Wilson 95\% $[0.0601, 0.1031]$, exceeding the preregistered floor of 40). Both control pairs pass C1 with substantial margins:
\begin{equation}
  \text{L23 pair: } 0.5628\ [0.3932, 0.7302], \qquad
  \text{L345 pair: } 0.8542\ [0.6088, 1.0995] \qquad (n = 48),
  \label{eq:c1}
\end{equation}
and the best site, L23$_{\obs}$, attains mean release rate
\begin{equation}
  \Rsite \;=\; 0.8886 \;\ge\; 0.5 \quad \text{(C2, pass)},
  \label{eq:c2}
\end{equation}
i.e.\ the transplant restores, on average, $89\%$ of the behavioral shift that evidence zeroing removes (Eq.~\eqref{eq:R}). A preregistered $\pm 0.01$ guard-band sensitivity check leaves both pairs' verdicts unchanged. Figure~\ref{fig:local} shows the pilot site screen and the confirmatory paired advantages; per protocol, pilot numbers are reported alongside confirmatory ones (Table~\ref{tab:local}).

\begin{table}[t]
\centering
\small
\caption{B0-rep localization ledger. C1 requires both paired advantages' 97.5\% CIs to exclude zero; C2 requires best-site release rate $\ge 0.5$. Sub-sample bookkeeping: fork $n{=}10$ ineligible worlds and IV $n{=}39$ are reported in the audit. Cost: $3.82$ GPU-hours.}
\label{tab:local}
\begin{tabular}{@{}llll@{}}
\toprule
Criterion & Statistic & Value & Verdict \\
\midrule
Eligibility & Eligible worlds $M$ & $48/608$ (Wilson $[0.0601, 0.1031]$) & \cellcolor{green!12}PASS \\
C1 & L23 paired advantage & $0.5628$ $[0.3932, 0.7302]$ & \cellcolor{green!12}PASS \\
C1 & L345 paired advantage & $0.8542$ $[0.6088, 1.0995]$ & \cellcolor{green!12}PASS \\
C2 & Best site (L23$_{\obs}$) mean $R$ & $0.8886$ & \cellcolor{green!12}PASS \\
Guard & $\pm 0.01$ sensitivity & both pairs still pass & \cellcolor{green!12}PASS \\
\bottomrule
\end{tabular}
\end{table}

\paragraph{What localization does and does not establish.} C1/C2 establish the preregistered claim G1 verbatim: \emph{``the suppression-by-default under OOD is localized to the observation-evidence row channels, and transplant intervention is sufficient to restore the target behavior.''} They do not establish that a \emph{deployable} pipeline can reproduce the transplant: the transplant uses oracle access to the released counterpart of each suppressed world. Converting localization into release requires a detector (to know when to intervene) and a generator (to construct the direction from the suppressed prompt alone). Those are the components that fail next.

\section{Failure 1: The Gate Inverts Out of Distribution}\label{sec:gate}

B6.5 is the preregistered terminal test of the full gated pipeline (Figure~\ref{fig:pipeline}): detector $u$ with threshold $T$, contrast pass, calibration-frozen direction $\delta_{\mathrm{ZERO}}$ (Eq.~\eqref{eq:dzero}), injection, guarded parse. Five criteria (B65a--e) were frozen before calibration; the verdict was bound to a claim template before any CONF50/ID40 data existed.

\paragraph{Calibration looked clean.} The detector is a light linear probe: PCA ($k{=}8$) over the site features $\phi$ (Eq.~\eqref{eq:phi}) followed by logistic regression (300 epochs, lr $10^{-2}$), labeled $y{=}1$ iff the world's denominator satisfies $D > 0.05$, with the per-seed threshold fixed as the intersection across seeds $\{0,1,2\}$. On the 40-world calibration set (24/8/8 split, 3 seeds), the detector's threshold settled at $T \in \{0.699, 0.728, 0.737\}$ with \emph{zero} validation ID triggers, OOD recall $0.5$, parse rate $1.0$, and direction norms $\lVert\delta\rVert_2 = 8.30$ (two-segment) / $14.35$ (three-segment). By every calibration-visible measure, the gate was conservative to the point of never firing.

\paragraph{The terminal test failed, and the failure was silent.} On held-out data the pipeline failed three of five criteria (Table~\ref{tab:b65}). The release intercept is $b = 0.3820$ (boot 95\% $[0.3396, 0.4230]$) against the frozen threshold $b \le 0.08$ (B65a, FAIL); ID trigger rates are $0.0688 / 0.0688 / 0.0734$ ($44/44/47$ of $n_{\mathrm{gen}} = 640$) against the frozen cap $0.02$ (B65e, FAIL); and the paired McNemar test is degenerate ($b{=}c{=}0$, $p{=}\mathrm{NaN}$) with $25/27$ negative-ATE worlds sign-reversed---against a cap of $14/27$ (B65c, FAIL). Two criteria passed: $|\Delta a| = 0.0$ with positive-world correlation $0.6299$ (B65b), and ID slope degradations of $0.0038$--$0.0042$ under the $0.05$ cap (B65d).

\begin{table}[t]
\centering
\small
\caption{B6.5 terminal criteria, all frozen before data. Bold rows failed. The gate triggered on $0/2{,}400$ CONF50 generations, so the ``gated'' pipeline is numerically identical to the base model on exactly the worlds the intervention exists for.}
\label{tab:b65}
\begin{tabular}{@{}lllll@{}}
\toprule
Criterion & Measurement & Threshold (frozen) & Observed & Verdict \\
\midrule
B65a & Release intercept $b$ (Eq.~\eqref{eq:fit}) & $\le +0.08$ & $\mathbf{0.3820}\ [0.3396, 0.4230]$ & \cellcolor{red!12}\textbf{FAIL} \\
B65b & Slope stability / corr & $|\Delta a| \le 0.10 \;\wedge\; \mathrm{corr} > 0.40$ & $0.0$; corr $0.6299$ ($n_{\mathrm{pos}}{=}23$) & \cellcolor{green!12}PASS \\
B65c & Reversals + McNemar & $\le 14/27 \;\wedge\; p < 0.05$ & $\mathbf{25/27}$; $b{=}c{=}0$, $p{=}\mathrm{NaN}$ & \cellcolor{red!12}\textbf{FAIL} \\
B65d & ID slope degradation & $\le 0.05$ & $0.0038$--$0.0042$ & \cellcolor{green!12}PASS \\
B65e & ID trigger rate & $\le 0.02$ & $\mathbf{0.0688}$--$\mathbf{0.0734}$ & \cellcolor{red!12}\textbf{FAIL} \\
\midrule
Gate audit & CONF50 triggers & (diagnostic) & $\mathbf{0/2{,}400}$ & --- \\
\bottomrule
\end{tabular}
\end{table}

\begin{figure}[t]
\centering
\includegraphics[width=\textwidth]{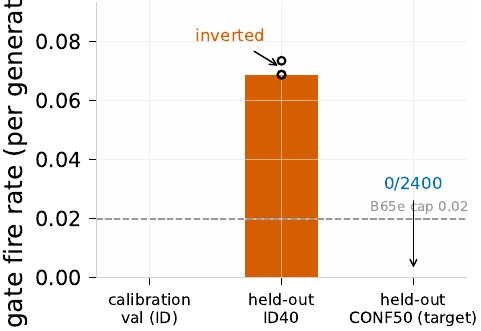}
\caption{The gate inversion. Detector trigger rates at calibration (left; $0.0$ on validation ID by construction), on held-out ID40 (center; $6.9$--$7.3\%$ across three seeds, above the frozen $0.02$ cap, dashed line), and on held-out CONF50 (right; $0/2{,}400$). The detector is simultaneously too permissive where intervention is harmful and absolutely silent where it is needed.}
\label{fig:gate}
\end{figure}

\paragraph{Why it failed: an OOD-inverted detector.} The diagnostic row of Table~\ref{tab:b65} is the finding. Across all $2{,}400$ CONF50 generations---the worlds for which the entire apparatus exists---the gate triggered \emph{zero} times: GATE $\equiv$ BASE, and the intervention never happened (Figure~\ref{fig:gate}). Simultaneously the same detector fired on $6.9$--$7.3\%$ of held-out ID generations where intervention is harmful. The detector learned ``looks like calibration OOD'' rather than ``needs intervention,'' and calibration, by construction, contained no example of the positive class's held-out realization. Crucially, \emph{no intermediate statistic raised an alarm}: calibration metrics were perfect, ID behavior was preserved (B65d), and slope stability passed (B65b). Without the preregistered gate audit, the failure would have read as ``the intervention does nothing''---exactly the misreading our own audit log initially recorded and then corrected on inspection of the per-generation records (Appendix~\ref{app:audit}).

\paragraph{Attribution is ambiguous---and that is a problem.} B6.5's frozen claim template states the disjunction faithfully: \emph{representation-level localization holds (B0-rep G1), while behavior-level release is not achieved at this resolution with this method.} But B6.5 alone cannot separate two causes: (i)~the gate never fired, so the release direction was never exercised at test time; (ii)~the direction itself might be insufficient even if exercised. The B6.6 amendment---archived before any B6.6 data---removes the gate entirely to decide between them.

\section{Failure 2: The Linear Release Family Is Bounded}\label{sec:release}

\subsection{The B6.6 amendment}

The B6.6 amendment (archived with hashes before any corresponding data) makes one surgical change: \emph{remove the gate}. The detector score $u$ plays no role; the per-instance direction $\delta(x)$ (Eq.~\eqref{eq:dx}) has zero learned parameters; injection (Eq.~\eqref{eq:inject}) is unconditional, at two doses $\lambda \in \{0.5\}$ (\textbf{ARM-S}) and $\lambda = 1.0$ (\textbf{ARM-R}), with an always-on fixed-direction control re-measured within the experiment (P1 of the amendment). BASE is re-measured in-experiment as well. The amendment froze a three-valued readout with mutually exclusive priority (P2): \textbf{R1} (a release margin exists $\to$ the pipeline design was right and only the gate failed), \textbf{R2} (no margin in either arm $\to$ the linear release family is bounded at this site and resolution), \textbf{R3} (intermediate), with INDETERMINATE and DESCRIPTIVE-ONLY fallbacks for insufficient ID power or unclassifiable patterns (P3--P5).

\subsection{Result: R2, no margin}

\begin{table}[t]
\centering
\small
\caption{B6.6 terminal readout (verdict \textbf{R2: no margin}). Dose--response is monotone in $\lambda$ for both intercept (down) and slope (down); neither arm approaches the frozen release threshold $b \le 0.08$; ID behavior is not harmed; per-instance adaptivity adds $\approx 0$ over the fixed direction. Cost: $0.1476$ GPU-hours.}
\label{tab:b66}
\begin{tabular}{@{}llll@{}}
\toprule
Quantity & BASE ($\lambda{=}0$) & ARM-S ($\lambda{=}0.5$) & ARM-R ($\lambda{=}1.0$) \\
\midrule
Intercept $b$ (Eq.~\eqref{eq:fit}) & $0.3820$ & $0.3106$ & $\mathbf{0.2637}$ \\
Slope $a$ & $0.8236$ & $0.6773$ & $0.5721$ \\
ID slope drop & --- & $-0.0243$ & $-0.0169$ \\
Reversals (neg-ATE worlds) & $25/27$ & --- & $21/27$ \\
McNemar (paired) & --- & --- & $b{=}1, c{=}1, p{=}0.75$ \\
\midrule
Per-world gain vs.\ always-on & --- & $+0.0268$ & $-0.0199$ \\
Positive-world corr (ARM-R) & \multicolumn{3}{l}{$0.328$} \\
\bottomrule
\end{tabular}
\end{table}

\begin{figure}[t]
\centering
\includegraphics[width=\textwidth]{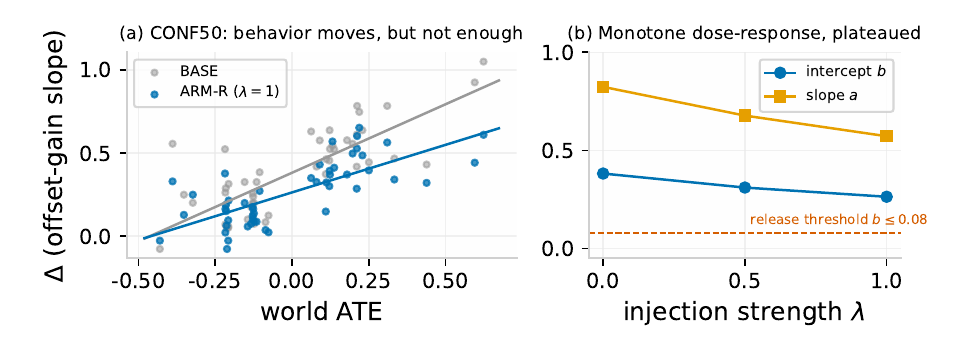}
\caption{B6.6 dose--response on CONF50. \textbf{(a)} Per-world offset-gain slope $\Delta$ (Eq.~\eqref{eq:delta}) vs.\ ground-truth ATE for BASE and ARM-R; injection shifts the regression toward the ideal diagonal but the fitted intercept (Eq.~\eqref{eq:fit}) remains far from zero. \textbf{(b)} Fitted intercept $b$ and slope $a$ as functions of dose $\lambda$; both are monotone, and the intercept plateaus at more than three times the frozen release threshold (dashed line).}
\label{fig:dose}
\end{figure}

\begin{figure}[t]
\centering
\includegraphics[width=0.5\textwidth]{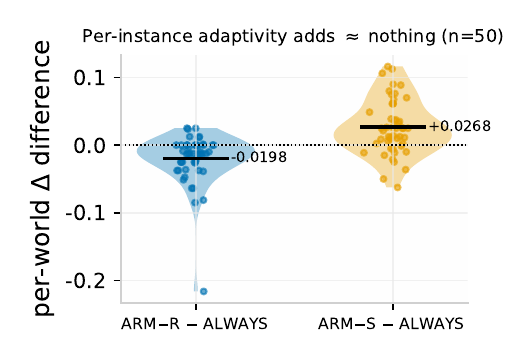}
\caption{Per-instance adaptivity contributes approximately nothing. Per-world difference between each adaptive arm and the always-on fixed direction ($n = 50$ worlds; black bars mark means $-0.0199$ and $+0.0268$). The remaining release gap after full-dose injection is $\approx 0.18$, an order of magnitude larger than any adaptive gain.}
\label{fig:adaptive}
\end{figure}

B6.6 (Table~\ref{tab:b66}) answers the attribution question left open by B6.5, and the answer is: \emph{both}. Unconditional injection moves behavior in the right direction---the intercept falls monotonically with dose, $0.3820 \to 0.3106 \to 0.2637$ (Figure~\ref{fig:dose})---but the full-strength arm remains more than $0.18$ above the frozen release threshold, and pushing dose further trades slope for intercept along the same monotone curve rather than closing the gap. The shrunk arm loses intercept ground monotonically, confirming a smooth dose--response rather than a threshold effect. ID behavior is unharmed in both arms (slope drops $-0.017$ and $-0.024$), so the bound is not an artifact of collateral damage. Per-instance adaptivity, the entire point of the $\delta(x)$ construction, contributes $-0.0199$ (ARM-R) and $+0.0268$ (ARM-S) per world over the always-on fixed direction (Figure~\ref{fig:adaptive}): statistically and practically nothing, against the $\approx 0.18$ gap that remains. The preregistered readout is therefore \textbf{R2\_NO\_MARGIN}, with its verbatim claim: \emph{per-instance linear directions at this site and resolution likewise cannot be released---the linear release family, fixed and adaptive alike, is bounded as a whole.}

\paragraph{Consequences frozen in advance.} Under the amendment's own frozen clauses, R2 prohibits the follow-up escalation experiment (B7) from launching: there is no release margin to route. The decision tree's terminal state is thus the conjunction we report as our main result---\emph{the gate is OOD-inverted} \emph{and} \emph{the linear release family at the localized site is bounded below the margin}, with localization intact upstream.

\section{Discussion}\label{sec:discussion}

\paragraph{The representation--behavior gap is not one gap.} Our three positive/negative pairs---decodable but not behaviorally coupled (stage 1), localizable but not gateable (B0-rep vs.\ B6.5), gate-removable but not linearly releasable (B6.5 vs.\ B6.6)---show that ``the model knows but doesn't use'' decomposes into at least three independent obstacles, each sufficient to sink a naive pipeline. Reporting any one of them alone would support a different (and wrong) remediation story: better probes (stage 1 already falsifies), better calibration (B6.6 falsifies), better sites (B0-rep falsifies). The value of the end-to-end design is that the failure stories are forced to be compatible with one another.

\paragraph{Silent gating failures are a distinct evaluation hazard.} B6.5's most dangerous property is that it is \emph{quiet}: every calibration statistic was perfect, ID behavior was preserved, and the intervention simply never ran on the worlds that mattered. Any gated or triggered intervention system---including selective steering, refusal circuits, and tool-routing heads---can fail in this mode, and standard held-out metrics will not catch it unless the trigger itself is audited on the deployment distribution. We recommend that gate-level trigger counts on the target distribution be reported as a first-class metric wherever a gate exists, and note that our own audit initially misread this failure before the per-generation records corrected it (Appendix~\ref{app:audit})---the misreading is the default; only instrumentation prevents it.

\paragraph{Bounded linearity at a verified site constrains steering practice.} Steering-vector methods report average behavioral shifts~\citep{turner2023activation,zou2023representation,li2023inference,subramani2022extracting}, and their unreliability under distribution shift has been documented~\citep{tan2024analysing}. B6.6 adds a sharper datum: at a site whose causal sufficiency was \emph{independently verified} by transplant, the per-instance linear family exhibits a smooth, monotone dose--response that plateaus at more than three times the required margin, with adaptivity worth less than $\pm 0.03$. If linear directions are insufficient even where localization is certain, then steering failures elsewhere cannot be automatically attributed to ``wrong layer'' or ``wrong direction'': the linear parametrization itself is a candidate bottleneck. Nonlinear or multi-site release mechanisms are the obvious next object of study---under the same preregistered discipline, since the space of post-hoc release mechanisms is large and forgiving.

\paragraph{Relation to concurrent work.} \citet{geometrysteering2026} document ``perfect detection, failed control'' in large models, convergent with our gate result. Our decomposition suggests their failed control may itself be composite (gate inversion, direction boundedness, or both), and that resolution-level reporting---trigger counts, dose--response curves, and per-world adaptive gains---can separate the causes in their setting too.

\paragraph{Limitations.} (i)~A single 25.7M model on a synthetic causal domain: the specific numbers are C1's, and only the \emph{dissociation pattern} is a candidate for generality. (ii)~The release family tested is linear, single-site, and residual-additive; our negative claim is explicitly scoped to that family at that resolution and does not touch nonlinear or distributed release. (iii)~The denominator guard excludes $39/50$ stage-1 worlds from rate statistics; rates are computed on measurable worlds only, and the audit reports the exclusions. (iv)~The eligibility rate for localization ($7.9\%$) means C1/C2 characterize a selected subpopulation of worlds; we report the selection rate and its interval rather than extrapolating. (v)~The gate was calibrated without held-out examples of the positive class's test-time realization; a detector trained with such examples might behave differently---but that is precisely the deployment condition the pipeline was designed for, and it failed there.

\paragraph{Why preregistration mattered here.} Every tempting post-hoc move was foreclosed in advance: re-thresholding the gate after seeing CONF50, redefining release around the slope rather than the intercept, promoting the adaptive-gain sign flip into a story, or launching the escalation experiment despite R2. The frozen branches forced the paper to be about the failure decomposition rather than about whichever sub-result could be made to look positive. We consider this the methodological contribution most likely to transfer.

\section{Conclusion}\label{sec:conclusion}

We subjected the detect--localize--release pipeline to a fully preregistered end-to-end test on a system with a documented representation--behavior gap. Localization passed decisively: the suppression is site-specific and transplant-sufficient. Release then failed twice, independently: the detector inverted out of distribution and silently disabled the entire intervention path, and the gate-free linear direction family plateaued at more than three times the preregistered threshold with negligible adaptive gain. Representation-level localization and behavior-level release are different problems, and the second is not solved by solving the first. For the growing literature that equates finding structure with controlling it, our results supply a concrete, audited counterexample---and a discipline (hashed criteria, verbatim claim templates, gate-level auditing, mandatory pilot reporting) for making the next counterexample harder to explain away, including by ourselves.

\section*{AI Use Statement}
A large-language-model assistant was used to help draft and copy-edit the manuscript, to adversarially proof-check its claims against the underlying experimental data and audit archive, and to develop and debug the evaluation code. All scientific claims, experimental design, data, analyses, and conclusions were made and verified by the author, who takes full responsibility for the content of this paper.

\section*{Ethics Statement}
This work studies a small synthetic-domain model and introduces no human-subjects, privacy, or dual-use concerns. Its negative results concern the limits of activation interventions and, if anything, counsel caution about intervention-based control claims. All data are generated by code released with the audit artifacts.

\section*{Reproducibility Statement}
Every number in this paper traces to a hashed artifact in the released audit chain: criteria files were MD5-archived before any corresponding data existed; code manifests, configurations, per-world records, and console logs are included; the model checkpoint is the frozen public artifact of \citet{xun2026evidence}; and all verdicts are reproducible by recomputing the archived statistics from the released records (we did so independently for both terminal experiments; the recomputation matched the pipeline's own analysis field-by-field). Appendix~\ref{app:audit} summarizes the chain.

\bibliographystyle{iclr2026_conference}
\bibliography{references}

\begin{thebibliography}{24}
\providecommand{\natexlab}[1]{#1}
\providecommand{\url}[1]{\texttt{#1}}
\expandafter\ifx\csname urlstyle\endcsname\relax
  \providecommand{\doi}[1]{doi: #1}\else
  \providecommand{\doi}{doi: \begingroup \urlstyle{rm}\Url}\fi

\bibitem[Anonymous(2026)]{geometrysteering2026}
Anonymous.
\newblock The geometry of knowing vs.\ steering: Perfect detection, failed
  control.
\newblock \emph{arXiv preprint arXiv:2606.24952}, 2026.

\bibitem[Arditi et~al.(2024)Arditi, Obeso, Syed, Paleka, Panickssery, Gurnee,
  and Nanda]{arditi2024refusal}
Andy Arditi, Oscar Obeso, Aaquib Syed, Daniel Paleka, Nina Panickssery, Wes
  Gurnee, and Neel Nanda.
\newblock Refusal in language models is mediated by a single direction.
\newblock \emph{Advances in Neural Information Processing Systems},
  37:\penalty0 136037--136083, 2024.

\bibitem[Belinkov(2022)]{belinkov2022probing}
Yonatan Belinkov.
\newblock Probing classifiers: Promises, shortcomings, and advances.
\newblock \emph{Computational Linguistics}, 48\penalty0 (1):\penalty0 207--219,
  2022.

\bibitem[Burns et~al.(2023)Burns, Ye, Klein, and
  Steinhardt]{burns2023discovering}
Collin Burns, Haotian Ye, Dan Klein, and Jacob Steinhardt.
\newblock Discovering latent knowledge in language models without supervision.
\newblock \emph{International Conference on Learning Representations}, 2023.

\bibitem[Efron \& Tibshirani(1993)Efron and Tibshirani]{efron1993introduction}
Bradley Efron and Robert~J. Tibshirani.
\newblock \emph{An Introduction to the Bootstrap}.
\newblock Chapman \& Hall, 1993.

\bibitem[Geiger et~al.(2022)Geiger, Wu, Lu, Rozner, Kreiss, Icard, Goodman, and
  Potts]{geiger2022inducing}
Atticus Geiger, Zhengxuan Wu, Hanson Lu, Josh Rozner, Elisa Kreiss, Thomas
  Icard, Noah Goodman, and Christopher Potts.
\newblock Inducing causal structure for interpretable neural networks.
\newblock \emph{International Conference on Machine Learning}, pp.\
  7324--7338, 2022.

\bibitem[Hewitt \& Liang(2019)Hewitt and Liang]{hewitt2019designing}
John Hewitt and Percy Liang.
\newblock Designing and interpreting probes with control tasks.
\newblock In \emph{Proceedings of the 2019 Conference on Empirical Methods in
  Natural Language Processing}, pp.\  2733--2743, 2019.

\bibitem[Jin et~al.(2023)Jin, Chen, Leeb, Gresele, Kamal, Lyu, Blin, Adauto,
  Kleiman-Weiner, Sachan, and Sch{\"o}lkopf]{jin2023cladder}
Zhijing Jin, Yuen Chen, Felix Leeb, Luigi Gresele, Ojas Kamal, Zhiheng Lyu,
  Kevin Blin, Fernando~Gonzalez Adauto, Max Kleiman-Weiner, Mrinmaya Sachan,
  and Bernhard Sch{\"o}lkopf.
\newblock Cladder: Assessing causal reasoning in language models.
\newblock \emph{Advances in Neural Information Processing Systems},
  36:\penalty0 31038--31065, 2023.

\bibitem[Kornblith et~al.(2019)Kornblith, Norouzi, Lee, and
  Hinton]{kornblith2019similarity}
Simon Kornblith, Mohammad Norouzi, Honglak Lee, and Geoffrey Hinton.
\newblock Similarity of neural network representations revisited.
\newblock In \emph{International Conference on Machine Learning}, pp.\
  3519--3529, 2019.

\bibitem[Li et~al.(2023)Li, Patel, Vi{\'e}gas, Pfister, and
  Wattenberg]{li2023inference}
Kenneth Li, Oam Patel, Fernanda Vi{\'e}gas, Hanspeter Pfister, and Martin
  Wattenberg.
\newblock Inference-time intervention: Eliciting truthful answers from a
  language model.
\newblock \emph{Advances in Neural Information Processing Systems},
  36:\penalty0 41451--41530, 2023.

\bibitem[McNemar(1947)]{mcnemar1947note}
Quinn McNemar.
\newblock Note on the sampling error of the difference between correlated
  proportions or percentages.
\newblock \emph{Psychometrika}, 12\penalty0 (2):\penalty0 153--157, 1947.

\bibitem[Meng et~al.(2022)Meng, Bau, Andonian, and Belinkov]{meng2022locating}
Kevin Meng, David Bau, Alex Andonian, and Yonatan Belinkov.
\newblock Locating and editing factual associations in {GPT}.
\newblock In \emph{Advances in Neural Information Processing Systems},
  volume~35, pp.\  17359--17372, 2022.

\bibitem[Nosek et~al.(2018)Nosek, Ebersole, DeHaven, and
  Mellor]{nosek2018preregistration}
Brian~A. Nosek, Charles~R. Ebersole, Alexander~C. DeHaven, and David~T. Mellor.
\newblock The preregistration revolution.
\newblock \emph{Proceedings of the National Academy of Sciences}, 115\penalty0
  (11):\penalty0 2600--2606, 2018.

\bibitem[Pearl(2009)]{pearl2009causality}
Judea Pearl.
\newblock \emph{Causality: Models, Reasoning, and Inference}.
\newblock Cambridge University Press, 2nd edition, 2009.

\bibitem[Rimsky et~al.(2023)Rimsky, Gabrieli, Schulz, Tong, Hubinger, and
  Turner]{rimsky2024steering}
Nina Rimsky, Nick Gabrieli, Julian Schulz, Meg Tong, Evan Hubinger, and
  Alexander~Matt Turner.
\newblock Steering {Llama} 2 via contrastive activation addition.
\newblock \emph{arXiv preprint arXiv:2312.06681}, 2023.

\bibitem[Simmons et~al.(2011)Simmons, Nelson, and Simonsohn]{simmons2011false}
Joseph~P. Simmons, Leif~D. Nelson, and Uri Simonsohn.
\newblock False-positive psychology: Undisclosed flexibility in data collection
  and analysis allows presenting anything as significant.
\newblock \emph{Psychological Science}, 22\penalty0 (11):\penalty0 1359--1366,
  2011.

\bibitem[Subramani et~al.(2022)Subramani, Suresh, and
  Peters]{subramani2022extracting}
Nishant Subramani, Niveditha Suresh, and Matthew~E. Peters.
\newblock Extracting latent steering vectors from pretrained language models.
\newblock In \emph{Findings of the Association for Computational Linguistics:
  ACL 2022}, pp.\  566--581, 2022.

\bibitem[Tan \& Chanin(2024)Tan and Chanin]{tan2024analysing}
Daniel~L. Tan and David Chanin.
\newblock Analysing the generalization and reliability of steering vectors.
\newblock \emph{Advances in Neural Information Processing Systems},
  37:\penalty0 139179--139202, 2024.

\bibitem[Turner et~al.(2023)Turner, Thiergart, Leech, Udell, Ulisse, Mini, and
  MacDiarmid]{turner2023activation}
Alexander~Matt Turner, Lisa Thiergart, Gavin Leech, David Udell, Juan Ulisse,
  Ni~Mini, and Monte MacDiarmid.
\newblock Activation addition: Steering language models without optimization.
\newblock \emph{arXiv preprint arXiv:2308.10248}, 2023.

\bibitem[Vaswani et~al.(2017)Vaswani, Shazeer, Parmar, Uszkoreit, Jones, Gomez,
  Kaiser, and Polosukhin]{vaswani2017attention}
Ashish Vaswani, Noam Shazeer, Niki Parmar, Jakob Uszkoreit, Llion Jones,
  Aidan~N. Gomez, {\L}ukasz Kaiser, and Illia Polosukhin.
\newblock Attention is all you need.
\newblock \emph{Advances in Neural Information Processing Systems}, 30, 2017.

\bibitem[Vig et~al.(2020)Vig, Gehrmann, Belinkov, Qian, Nevo, Singer, and
  Shieber]{vig2020investigating}
Jesse Vig, Sebastian Gehrmann, Yonatan Belinkov, Yaron Qian, Daniel Nevo, Yaron
  Singer, and Stuart Shieber.
\newblock Investigating gender bias in language models using causal mediation
  analysis.
\newblock \emph{Advances in Neural Information Processing Systems},
  33:\penalty0 12388--12401, 2020.

\bibitem[Wilson(1927)]{wilson1927probable}
Edwin~B. Wilson.
\newblock Probable inference, the law of succession, and statistical inference.
\newblock \emph{Journal of the American Statistical Association}, 22\penalty0
  (158):\penalty0 209--212, 1927.

\bibitem[Xun(2026)]{xun2026evidence}
Xining Xun.
\newblock Evidence-type competition: Suppression and release of latent causal
  structure in a small transformer.
\newblock \emph{arXiv preprint arXiv:2607.29484}, 2026.

\bibitem[Zou et~al.(2023)Zou, Phan, Chen, Campbell, Guo, Ren, Pan, Yin,
  Mazeika, Dombrowski, Goel, Li, Byun, Wang, Mallen, Basart, Koyejo, Song,
  Fredrikson, Kolter, and Hendrycks]{zou2023representation}
Andy Zou, Long Phan, Sarah Chen, James Campbell, Phillip Guo, Richard Ren,
  Alexander Pan, Xuwang Yin, Mantas Mazeika, Ann-Kathrin Dombrowski, Shashwat
  Goel, Nathaniel Li, Michael~J. Byun, Zifan Wang, Alex Mallen, Steven Basart,
  Sanmi Koyejo, Dawn Song, Matt Fredrikson, J.~Zico Kolter, and Dan Hendrycks.
\newblock Representation engineering: A top-down approach to {AI} transparency.
\newblock \emph{arXiv preprint arXiv:2310.01405}, 2023.

\end{thebibliography}

\appendix
\section{Audit Chain and Preregistered Claim Templates}\label{app:audit}

\subsection{Audit timeline}

Table~\ref{tab:audit} summarizes the append-only audit log (25 registry entries, B0-REG through B0-REG-19 including sub-entries). Each entry binds a criteria/code artifact by MD5 \emph{before} any data it governs; verdict entries additionally archive the result artifacts and an independent recomputation cross-check.

\begin{table}[h]
\centering
\footnotesize
\caption{Audit chain milestones. ``Hash-first'' means the artifact hash was archived before any governed data existed. Full log released with the artifacts.}
\label{tab:audit}
\begin{tabular}{@{}l>{\raggedright\arraybackslash}p{7.2cm}l@{}}
\toprule
Entry & Content & Discipline \\
\midrule
B0-REG-1$\dots$14 & Stage-1 gates B0/B0.5/B1; site selection; localization criteria C1/C2 & hash-first \\
B0-rep ledger & $M{=}48/608$; C1 pairs; C2 $0.8886$; pilot side-by-side & frozen claim G1 \\
B0-REG-15 & B6.5 calibration ledger (3 seeds, hashes of $u$, $\delta_{\mathrm{ZERO}}$) & hash-first \\
B0-REG-16 & B6.5 verdict FAILED; gate $0/2{,}400$ audit; \textbf{authorial correction} (see below) & frozen template \\
B0-REG-17 & B6.6 amendment registered & before any B6.6 data \\
B0-REG-18 & B6.6 code + config hashes; P1--P5 pins; 31/31 self-tests & hash-first \\
B0-REG-19 & B6.6 verdict R2\_NO\_MARGIN; recomputation matched; B7 not launched & frozen template \\
\bottomrule
\end{tabular}
\end{table}

\paragraph{Authorial correction (B0-REG-16 addendum).} The first verbal reading of B6.5 recorded the failure as ``the injection perturbs behavior without benefit.'' Inspection of the per-generation records showed this was wrong: the gate triggered $0/2{,}400$ times, GATE $\equiv$ BASE, and the $25/27$ reversals are the \emph{base model's own} sign errors on negative-ATE worlds---the suppression phenomenon of \citet{xun2026evidence} reproducing itself, not an injection artifact. The correction is archived in the same log; we report it in the paper because it is exactly the misreading that un-instrumented gated systems invite (Section~\ref{sec:discussion}).

\paragraph{Registered code imperfection.} The B6.5 records writer hard-coded \texttt{"partial": true}; the field is cosmetic, was registered rather than silently patched (the code was frozen), and the B6.6 writer was corrected to terminal-state flipping (verified: B6.6 records show \texttt{partial=false}).

\subsection{Verbatim claim templates}

The frozen templates governing the terminal verdicts. The canonical archived text is Chinese; the English renderings below are faithful translations prepared for this paper, with the Chinese originals included in the released audit log.

\begin{tcolorbox}[colback=gray!4, colframe=gray!50, title={B6.5 failure template (invoked)}, fonttitle=\footnotesize\bfseries, left=2mm, right=2mm, boxrule=0.4pt]
\footnotesize
Representation-level localization holds (B0-rep G1), while behavior-level release is not achieved at this resolution with this method.
\end{tcolorbox}

\begin{tcolorbox}[colback=gray!4, colframe=gray!50, title={B0-rep G1 template (invoked)}, fonttitle=\footnotesize\bfseries, left=2mm, right=2mm, boxrule=0.4pt]
\footnotesize
The suppression-by-default under OOD is localized to the observation-evidence row channels, and transplant intervention is sufficient to restore the target behavior.
\end{tcolorbox}

\begin{tcolorbox}[colback=gray!4, colframe=gray!50, title={B6.6 R2 template (invoked)}, fonttitle=\footnotesize\bfseries, left=2mm, right=2mm, boxrule=0.4pt]
\footnotesize
Per-instance linear directions at this site and resolution likewise cannot be released---the linear release family, fixed and adaptive alike, is bounded as a whole.
\end{tcolorbox}

\subsection{Compute and artifacts}
Terminal-experiment costs: localization $3.82$ GPU$\cdot$h; B6.5 calibration $0.04$; B6.5 terminal $0.2125$; B6.6 $0.1476$; total $4.22$ GPU$\cdot$h. Released artifacts include the audit log, criteria and configuration files with hashes, code manifests (39 entries), per-world record files for both terminal experiments, console logs, and the figure-generation scripts consuming those records.

\end{document}